\documentclass{article} % For LaTeX2e
\usepackage{iclr2023_conference,times}

\usepackage{amsmath,amsfonts,bm}

\def\eqref#1{equation~\ref{#1}}
\def\1{\bm{1}}

\DeclareMathAlphabet{\mathsfit}{\encodingdefault}{\sfdefault}{m}{sl}
\SetMathAlphabet{\mathsfit}{bold}{\encodingdefault}{\sfdefault}{bx}{n}

\newcommand{\E}{\mathbb{E}}

\newcommand{\KL}{D_{\mathrm{KL}}}

\DeclareMathOperator*{\argmax}{arg\,max}

\usepackage{hyperref}
\usepackage{url}
\usepackage{amssymb,graphicx,mathtools,amsmath}
\usepackage{multirow,booktabs}
\usepackage{caption,subcaption,float,comment}

\DeclareMathOperator{\x}{\mathbf{x}}
\DeclareMathOperator{\thet}{\boldsymbol{\theta}}

\title{Stabilized training of joint energy-based models and their practical applications
}

\author{Martin Sustek \\ 
Faculty of Information Technology, Brno University of Technology, Czechia - FIT BUT \\
Center for Language and Speech Processing, Johns Hopkins University, USA - CLSP JHU \\
\texttt{isustek@fit.vut.cz}
\And
Samik Sadhu \\
CLSP JHU
\And
Lukas Burget \\
FIT BUT \\
\And
Hynek Hermansky \\
CLSP JHU \\
HLTCOE, JHU, USA
\AND
Jesus Villalba \\
CLSP JHU \\
HLTCOE, JHU, USA
\And
Laureano Moro-Velazquez \\
CLSP JHU
\And
Najim Dehak \\
CLSP JHU \\
HLTCOE, JHU, USA
\AND
}

\iclrfinalcopy % Uncomment for camera-ready version, but NOT for submission.
\begin{document}

\maketitle

%\DeclareMathOperator*{\argmax}{arg\,max}
%\DeclareMathOperator*{\argmin}{arg\,min}
%\DeclareMathOperator{\E}{\mathbb{E}}

% Template for ICASSP-2021 paper; to be used with:
%          spconf.sty  - ICASSP/ICIP LaTeX style file, and
%          IEEEbib.bst - IEEE bibliography style file.
% --------------------------------------------------------------------------

%
\begin{abstract}
The recently proposed Joint Energy-based Model (JEM) interprets discriminatively trained classifier $p(y|x)$ as an energy model, which is also trained as a generative model describing the distribution of the input observations $p(x)$. The JEM training relies on "positive examples" (i.e. examples from the training data set) as well as on "negative examples", which are samples from the modeled distribution $p(x)$ generated by means of Stochastic Gradient Langevin Dynamics (SGLD). Unfortunately, SGLD often fails to deliver negative samples of sufficient quality during the standard JEM training, which causes a very unbalanced contribution from the positive and negative examples when calculating gradients for JEM updates. As a consequence, the standard JEM training is quite unstable requiring careful tuning of hyper-parameters and frequent restarts when the training starts diverging. This makes it difficult to apply JEM to different neural network architectures, modalities, and tasks. In this work, we propose a training procedure that stabilizes SGLD-based JEM training (ST-JEM) by balancing the contribution from the positive and negative examples. We also propose to add an additional "regularization" term to the training objective -- MI between the input observations $x$ and output labels $y$ -- which encourages the JEM classifier to make more certain decisions about output labels. We demonstrate the effectiveness of our approach on the CIFAR10 and CIFAR100 tasks. We also consider the task of classifying phonemes in a speech signal, for which we were not able to train JEM without the proposed stabilization. We show that a convincing speech can be generated from the trained model. Alternatively, corrupted speech can be de-noised by bringing it closer to the modeled speech distribution using a few SGLD iterations. We also propose and discuss additional applications of the trained model.
%Joint Energy-based Model (JEM) introduced a framework that unifies discriminative and generative model. JEM is trained using samples obtained by Stochastic Gradient Langevin Dynamics (SGLD) and its training usually diverges which makes it difficult to apply JEM to a different modality, model or task. In this work, we make an effort toward optimizing MI between inputs and labels and introduce a stable way of training with SGLD samples. We also inspect SGLD procedure and make changes to it based on our findings. This allows us to apply JEM on speech as a phoneme classifier where we demonstrate advantages and possible future work. We also demonstrate the capability of model to produce speech features.
\end{abstract}

%
%\begin{keywords}
%One, two, three, four, five
%\end{keywords}
%
\section{Introduction}\label{sec:intro}
One of the most common Machine Learning tasks is to classify input data points into chosen categories which typically is accomplished by a discriminatively trained classifier. Having an enormous amount of data and a powerful machine learning model of suitable architecture with a huge number of learnable parameters is usually enough to get close to state-of-the-art performance. The alternative approach of training a generative model and then inferring the posterior probability over the possible categories can outperform discriminative models only in the low-resource settings and usually fails to be competitive due to its restricted modeling power otherwise. The utility of explicit generative models then lies in the access to the likelihood of the input data useful for e.g. detecting outliers while implicit generative models are evaluated based on the capability to generate realistic and diverse input, especially when applied to images. The promise of generative models to avoid expensive labeling of unlabeled data (for which a discriminative classifier has no use) is shadowed by the recent success of self-supervised techniques that take advantage of self-contained information in the time sequence or leverage known input data manipulations (e.g. shift and resize of images) to introduce labels used to train a discriminative model. Self-supervised models are usually used as pre-trained models to extract embeddings that work well in the downstream task. 

Recently, \cite{grathwohl2019your} showed that every discriminatively trained classifier can be seen as an energy-based model modeling the joint distribution between the input data $x$ and the label $y$. In fact, when the discriminator is trained in a standard way, we are only training the model to provide us with a good estimate of $p(y|x)$ while $p(x)$ is not being optimized at all. In order to optimize $p(x)$, authors used Stochastic Gradient Langevin Dynamics (SGLD) as it was described in \cite{welling2011bayesian} to sample\footnote{These samples are sometimes called negative samples as opposed to train data being called positive samples.} from the modeled distribution and called the resulting model Joint Energy-based Model (JEM). Authors demonstrated that the longer training time and slight performance degradation (compared to its strictly discriminative counterpart) is compensated by the possibility to generate either category-conditional or unconditional samples, robustness against adversarial attacks, improved calibration, and out-of-distribution detection. Unfortunately, it is difficult to take these models and easily apply them to new tasks as the training often diverges, and restarting the training from the last saved epoch seems to be the only reliable solution to reach a stable optimum. This prevents the community from conducting a deeper exploration of these models. As a response, VERA (\cite{duvenaud2021no}) introduced a stable way of training JEMs. In this work, the authors demonstrate that they are able to speed up the training by introducing an auxiliary model (generator) and they are capable of producing high-quality images based on that generator. 

\section{Our contribution} \label{sec:contrib}
In this work, we propose a method to train JEM without the use of any auxiliary model whose training does not diverge even though we use SGLD samples. We propose to add a term that maximizes mutual information (MI) between the inputs $x$ and labels $y$ into the previously defined objective function and approximate it for necessary simplification. This reveals an alternative way of training and allows us to reach superior accuracy compared to results reported by \cite{grathwohl2019your} using JEMs on the CIFAR-10 and CIFAR-100 dataset while being able to generate images of similar quality. 

Our method allows for training even in the case of not using any SGLD samples. In fact, this can be interpreted as optimizing JEM with the domain defined only over the training data points. {We cannot sample from the model as we assume that outside of the domain of train data, $p(x)$ is strictly $0$ and the model output is ignored.}. Over this restricted domain, we don't need to approximate MI but we can optimize it directly. This results in a boost in performance, explaining why our JEM model outperforms the baseline which was not the 
the case for the original training of JEM.

Maximizing approximate mutual information term added to the loss function helped us realize the stabilization trick, but we show that the same trick can also be directly applied to the joint distribution $p(x,y)$ (Excluding MI) reaching very similar performance.

\cite{grathwohl2019your} demonstrated that JEM can provide better calibration in the low-resource setting. They used only a portion of the original labeled data and treated the rest of the data as unlabeled. Unfortunately, they reported that adding unlabeled data had no effect on the accuracy. We show that low-resource or more difficult problems are the ones where we see the largest boost in accuracy and calibration compared to a simple discriminative model. Moreover, we extended our approach to be able to handle unlabeled data and we actually do report a boost in accuracy.

The updated version of SGLD has fixed hyperparameters. By carefully monitoring the SGLD process during the training, we noticed that the optimal hyper-parameters change during the training. Using the same hyper-parameters can eventually lead to the state when the updated version of SGLD does not provide any reasonable samples. In our approach, SGLD samples that are not competitive, do not influence the stability of the training. It opens the possibility to do a simple exploration of the different hyper-parameters, resulting in competitive samples again. This is a promising way to improve the speed and the quality of generated samples and more sophisticated ways should be explored in the future. We also found out that the quality and the speed of generated samples can be greatly affected by these and for conditional sampling, each class might have different optimal hyper-parameters.

Our main motivation to stabilize JEM training is to be able to apply it to a different modality (speech). We train JEM to model $p(x,y)$ of the input frame and its context\footnote{We used 80 log Mel-filter banks as a frame representation.} $x$ and the phoneme label of the central frame. We discuss the potential future use of JEM as a single model capable of being ASR, TTS, denoiser, speaker recognizer, voice conversion, the source separator or inpainting (also inpainting conditional on the category). The generative part of JEM can further be leveraged when estimating uncertainty [ignore the uncertain part of the input for ASR, SPK-ID] or when combining the output of different models. Furthermore, we demonstrate that we are able to generate interesting conditional and unconditional speech, and show promising preliminary results on denoising and model combination.

Our approach allows us to include any number of samples from SGLD and we typically use 8 samples per each batch of 64{\footnote{100 for speech experiments}} training examples. Increasing the number of samples increases the quality of generated images but at the same time slightly degrades the performance of a classifier. We have also observed an increase in the quality of generated images (CIFAR-10, CIFAR-100) during the training usually corresponds to a decrease in accuracy, this suggests that the model might not be large enough as we haven't noticed the same behavior when training on the speech datasets. An alternative way of speeding up the training is generating more SGLD samples but doing so only once per few batches. We have observed that increasing the number of SGLD steps can results in a much better quality of generated images in both training and 
inference but it significantly slows down the training. 
[When finished, add references from this section to the parts in the body/appendix where it is described with more details]

\section{Background}\label{sec:background}
In this section, we provide an overview and explanation of previously introduced techniques that helps to follow our reasoning. More detailed explanations of these techniques can be found in referenced literature.
\subsection{Energy-based models}
Energy-based models can represent a complex probability distribution. This is a special case of distributions because there is no simple way to sample from such a distribution - we can only evaluate probability distribution up to an unknown normalizer. Probability distribution over a continuous variable $\x$ is then defined as
\begin{equation} \label{eq:energy_px}
p_{\boldsymbol{\theta}}(\x) = \frac{e^{-E_{\boldsymbol{\theta}}(\x)}}{Z_{\boldsymbol{\theta}}} = \frac{e^{f_{\boldsymbol{\theta}}^x(\x)}}{Z_{\boldsymbol{\theta}}}\text{,}
\end{equation}
where $E_{\boldsymbol{\theta}}(\x)$ is the energy function that assigns a score to each continuous input $\x \in \mathbf{X}, \mathbf{X} = \mathbb{R}^{D_x}$. To make sure that $p_{\boldsymbol{\theta}}(\x)$ is a properly normalized distribution, we define a partition function as $Z_{\boldsymbol{\theta}}=\int_{\mathbf{x}}{e^{f_{\boldsymbol{\theta}}^x(\mathbf{x})}} d\mathbf{x}$. We can further modify Equation \ref{eq:energy_px} to define the joint distribution $p_{\boldsymbol{\theta}}(\mathbf{x},y)$ of both continuous input $\mathbf{x}$ and a discrete label $y \in \text{Y}, \text{Y} = \mathbb{N}^+, y \le D_y$ as
\begin{equation} \label{eq:energy_pxy}
p_{\boldsymbol{\theta}}(\mathbf{x},y=i) = \frac{e^{-E_{\boldsymbol{\theta}}(\mathbf{x},y=i)}}{Z_{\boldsymbol{\theta}}} = \frac{e^{f_{\boldsymbol{\theta}}(\mathbf{x},y=i)}}{Z_{\boldsymbol{\theta}}} = \frac{e^{{f_{\boldsymbol{\theta}}(\mathbf{x})}_{i}}}{Z_{\boldsymbol{\theta}}} = \frac{e^{q_i}}{Z_{\boldsymbol{\theta}}}
\end{equation}
A typical way of obtaining the value of negative energy $-E_{\boldsymbol{\theta}}(\mathbf{x},y)$ is via a function ${f_{\boldsymbol{\theta}}(\mathbf{x},y): \x ,y \mapsto -E_{\boldsymbol{\theta}}(\mathbf{x},y)}$, but we focus on an alternative way by using a vector-valued function $f_{\boldsymbol{\theta}}(\mathbf{x}): \x \mapsto \mathbf{q}$, $\mathbf{q} \in \mathbb{R}^{D_y}$, where the $i$-th element of the vector $\mathbf{q}$ (denoted as $q_i$) represents $-E_{\boldsymbol{\theta}}(\mathbf{x},y=i)$. In both cases, the partition function is given by $Z_{\boldsymbol{\theta}}=\sum_y\int_{\mathbf{x}}{e^{-E_{\boldsymbol{\theta}}(\mathbf{x},y)}}d\mathbf{x}$.   
Notice that EBMs do not provide an access to likelihood values because $Z_{\boldsymbol{\theta}}$ is intractable. Maximizing the log-likelihood of one data point $\mathbf{x}$ with respect to the parameters ${\boldsymbol{\theta}}$ is not straightforward either. In order to compute the gradient, we need to evaluate the intractable expectation over $\x$ as shown in Equation \ref{eq:grad_px}.
\begin{equation} \label{eq:grad_px}
\nabla_{\boldsymbol{\theta}} \log p_{\boldsymbol{\theta}}(\mathbf{x}) = \nabla_{\boldsymbol{\theta}} f_{\boldsymbol{\theta}}^x(\mathbf{x}) - \nabla_{\boldsymbol{\theta}} \log Z_{\boldsymbol{\theta}} =  \nabla_{\boldsymbol{\theta}} f_{\boldsymbol{\theta}}^x(\mathbf{x}) -  \E_{(\tilde{\mathbf{x}}) \sim p_{\boldsymbol{\theta}}(\mathbf{x})} \left[  \nabla_{\boldsymbol{\theta}} f_{\boldsymbol{\theta}}^x(\tilde{\mathbf{x}}) \right]
\end{equation}
Likewise, expressing $\nabla_{\boldsymbol{\theta}} \log p_{\boldsymbol{\theta}}(\mathbf{x},y)$ leads to the same conclusion (Equation \ref{eq:grad_pxy}).
\begin{equation} \label{eq:grad_pxy}
\nabla_{\boldsymbol{\theta}} \log p_{\boldsymbol{\theta}}(\mathbf{x},y) = \nabla_{\boldsymbol{\theta}} f_{\boldsymbol{\theta}}(\mathbf{x})_y - \nabla_{\boldsymbol{\theta}} \log Z_{\boldsymbol{\theta}} =  \nabla_{\boldsymbol{\theta}} f_{\boldsymbol{\theta}}(\mathbf{x})_y -  \E_{(\tilde{\mathbf{x}},j) \sim p_{\boldsymbol{\theta}}(\mathbf{x},y)} \left[  \nabla_{\boldsymbol{\theta}} f_{\boldsymbol{\theta}}(\tilde{\mathbf{x}})_j \right]
\end{equation}
One popular strategy to approximate these expectations is to generate samples from the modeled distribution, the ones based on Markov chain Monte Carlo (MCMC) are of particular interest to us, specifically, Stochastic Gradient Langevin Dynamics (SGLD) \cite{welling2011bayesian}\footnote{An overview of alternative approaches to train EBMs is presented in \cite{song2021train}.}. SGLD is capable of generating samples from $p_{\boldsymbol{\theta}}(\mathbf{x})$ induced by a neural network $f_{\boldsymbol{\theta}}^x(\mathbf{x})$ through an iterative procedure
\begin{equation} \label{eq:sgld}
\mathbf{x}^{t+1} = \x ^{t} + \frac{\alpha^t}{2} \nabla_{\mathbf{x}} f_{\boldsymbol{\theta}}^x(\mathbf{x}^{t}) + \mathbf{u}^t,\;\;\;\;\;\;\;\;\; u^t_i \sim \mathcal{N}(0,\alpha^t), \;\; 1 \le i \le D_x\text{,}
\end{equation}
starting from a random input $\x^0$.
In theory, when $\sum_{t} \alpha^t = \infty$ and $\sum_{t} (\alpha^t)^2 < \infty$,  it is guaranteed that $\mathbf{x}^t$ becomes a sample from  $p_{\boldsymbol{\theta}}(\mathbf{x})$ as $t \to \infty$. In practice, we must resort to an updated (sped up) version of SGLD by significantly limiting the number of steps. The number of steps needed to generate a reasonable sample can be even smaller when the initial sample $\x^0$ is chosen carefully -- one of the common techniques is called Persistent Contrastive Divergence (PCD), where the buffer of previously generated samples is maintained and initial samples are drawn from the buffer most of the time as described in \cite{du2019implicit}. Another trick is to reduce the amount of noise $\mathbf{u}^t$ added at each step. Notice, that in order to obtain a sample from $p_{\boldsymbol{\theta}}(\mathbf{x})$ when modeling the joint distribution $p_{\thet}(\x ,y)$, we need to replace $f_{\thet}^x(\x)$ by $\log \sum_i e^{f_{\boldsymbol{\theta}}(\mathbf{x})_i}$ in Equation \ref{eq:sgld}. If we desire to get a sample having a particular label $y=i$\footnote{This is a sample from both $p_{\thet}(\x,y=i)$ and $p_{\thet}(\x \mid y=i)$. $\log p_{\thet}(y)$ is a constant, so $\nabla_{\mathbf{x}}\log p_{\thet}(y) = 0$.}, $f_{\thet}^x(\x)$ is replaced by $f_{\boldsymbol{\theta}}(\mathbf{x})_i$. 

\subsection{Classification}
\label{sec:classification}
Training a classifier parameterized by ${\boldsymbol{\theta}}$ is achieved by minimizing the cross-entropy $\operatorname{H}$ between the true posterior distribution $p$ and modelled posterior distribution $p_{\boldsymbol{\theta}}$ over the possible values of the label $y$. For a given single input data point $\mathbf{x}$, we calculate 
\begin{equation} \label{eq:}
H(p, p_{\thet}) = -\sum_{y\in \text{Y}} p(y \mid \x) \log p_{\thet}(y \mid \x)\text{.}
\end{equation} 
We can directly access the true posterior distribution $p(y \mid \x)$ but $p(x)$ is inaccessible, for that reason we approximate the expectation over the true data distribution $\E_{p(\x)} \left[ \operatorname{H}(p,p_{\boldsymbol{\theta}})\right]$ by an empirical distribution $p_{data}$ that is provided in form of tuples $(\mathbf{x^i},y^i) \in \operatorname{DS}$, where $p_{data}(y=y^i \mid \mathbf{x}=\mathbf{x^i})=1$, therefore
\begin{equation} \label{eq:discriminative}
\E_{p_{\mathbf{x}}} \left[ \operatorname{H}(p,p_{\boldsymbol{\theta}})\right] \approx  \underbrace{\E_{\mathbf{x} \sim p_{data}(x)} \E_{y \sim p(y \mid \x )}}_{\E_{(\mathbf{x},y) \sim p_{data}(\mathbf{x},y)}} \left[-\log{p_{\boldsymbol{\theta}}(y \mid \x )}\right]  = \frac{1}{\lvert \operatorname{DS}\rvert} \sum_{i=0}^{\lvert \operatorname{DS}\rvert} -\log{p_{\boldsymbol{\theta}}(y^i \mid \mathbf{x^i})}
\end{equation}
In order to model $p_{\boldsymbol{\theta}}(y \mid \x )$, it is common to use Softmax function $\operatorname{SM(\mathbf{z})}: \mathbb{R}^{D_y} \to (0,1)^{D_y}$ (Equation \ref{eq:softmax} that transforms logits $\mathbf{z} = g_{\thet}(\x)$ into the vector of posterior probabilities, where $g_{\thet}(\x)$ is usually a neural network. 
\begin{equation} \label{eq:softmax}
p_{\boldsymbol{\theta}}(y=i \mid \x ) = \operatorname{SM}(\mathbf{z})_i = \frac{e^{z_i}}{\sum_{j=1}^{D_y} e^{z_j}}
\end{equation}
Softmax has one extra degree of freedom, meaning that for each $\x$, we can add any value $c(\mathbf{x})$ to logits $\mathbf{z}$ and it has no effect on the posterior distribution. Notice that for chosen $\x$, this value must be the same for all $y$, therefore we can obtain this value by any function $c(\mathbf{x}): \mathbb{R}^{D_x} \to \mathbb{R}$:
\begin{equation} \label{eq:sm_add_cx}
\operatorname{SM}_i(\mathbf{z}+c(\mathbf{x})) = \frac{e^{z_i+c(\mathbf{x})}}{\sum_{j=1}^{D_y} e^{z_j+c(\mathbf{x})}} = \frac{e^{c(\mathbf{x})}e^{z_i}}{e^{c(\mathbf{x})} \sum_{j=1}^{D_y} e^{z_j}} = \operatorname{SM}_i(\mathbf{z})
\end{equation}
From Equations \ref{eq:softmax} and \ref{eq:sm_add_cx} (replacing $e^{c(\mathbf{x})}$ by $k(\mathbf{x})$ for brevity), we can see:
\begin{equation}\label{eq:p_y_giv_x_clas}
p_{\boldsymbol{\theta}}(y=i \mid \x ) = \frac{e^{c(\mathbf{x})}e^{z_i}}{e^{c(\mathbf{x})} \sum_{j=1}^{D_y} e^{z_j}} = \frac{k(\mathbf{x})e^{z_i}}{k(\mathbf{x})\sum_{j=1}^{D_y} e^{z_j}} =  \frac{\frac{e^{z_i}}{k(\mathbf{x})}}{\sum_{j=1}^{D_y}\frac{e^{z_j}}{k(\mathbf{x})}}
\end{equation}

\subsection{Energy-based Classifier}
Expressing the posterior distribution $p_{\boldsymbol{\theta}}(y \mid x)$ of EBM by using product rule, applying sum rule and plugging into Equation \ref{eq:energy_pxy}, we have:
\begin{equation} \label{eq:p_y_giv_x_ebm}
p_{\boldsymbol{\theta}}(y=i \mid \x ) = \frac{p_{\boldsymbol{\theta}}(\mathbf{x},y=i)}{p_{\boldsymbol{\theta}}(\mathbf{x})} = \frac{p_{\boldsymbol{\theta}}(\mathbf{x},y=i)}{\sum_y p_{\boldsymbol{\theta}}(\mathbf{x},y=i)} = \frac{\frac{e^{q_i}}{Z_{\boldsymbol{\theta}}}}{\sum_{j=1}^{D_y}\frac{e^{q_j}}{Z_{\boldsymbol{\theta}}}}
\end{equation}
Comparing Equation \ref{eq:p_y_giv_x_clas} and Equation \ref{eq:p_y_giv_x_ebm}, we can observe that $\mathbf{z} = \mathbf{q}$ if we force logits obtained by discriminative model not to have $k(\x)$ dependent on $\x$ as ${Z_{\boldsymbol{\theta}}}$ is just a constant which is the same for every $\x$ and $y$.
This was observed by \cite{grathwohl2019your} and the model introduced in Equation \ref{eq:energy_pxy} is called the Joint Energy-based Model (JEM). They decided to model the joint distribution $\log p_{\boldsymbol{\theta}}(\mathbf{x},y)$ via the decomposition ${\log p_{\boldsymbol{\theta}}(\mathbf{x}) + \log p_{\boldsymbol{\theta}}(y\mid\mathbf{x})}$. This factorization is motivated by the fact that the updated (practical) version of SGLD (Equation \ref{eq:sgld}) cannot generate proper samples and this results in a biased gradient estimator of $p_{\thet}(\x)$ or $p_{\thet}(\x,y)$. This factorization enables training of an unbiased classifier, because $p_{\boldsymbol{\theta}}(y \mid \x )$ is not affected by generated samples and is trained in the standard way (Equation \ref{eq:softmax}). Generative part of this model is trained by maximizing $p_{\boldsymbol{\theta}}(\mathbf{x})$ as $\sum_y p_{\boldsymbol{\theta}}(\mathbf{x},y)$. 

The drawback of this approach is that producing samples by SGLD (Equation \ref{eq:sgld}) is time-consuming even if we resort to its updated version. Moreover, speeding up the sampling process by reducing the number of steps $t$ required to produce a sample causes their training to diverge \cite{grathwohl2019your}. In fact, the authors declared that the training instability was the most significant flaw of the proposed model. To resolve these instability issues, multiple approaches have been proposed, such as bypassing SGLD during the training \cite{grathwohl2020learning,duvenaud2021no}, applying SGLD only to fine-tune samples produced by a different generator \cite{xie2018cooperative}, applying SGLD in lower-dimensional hidden space \cite{che2020your} or restricting the model by adding spectral normalization to each layer and regularizing the energy of both generated and real samples \cite{du2019implicit}. 

%We provide more detailed and technical overview of previous work mentioned in the \ref{sec:intro} Appendix \ref{sec:background}. 
\section{Stabilized JEM}
We noticed that the energy of generated images by SGLD is not always comparable to the energy of the training data. Our idea to stabilize the training is guided by the observation that maximization of $\log p(y \mid \x)$ using Softmax function (Equation \ref{eq:softmax}) does not diverge during the training while EBM training using Equation \ref{eq:grad_px} or Equation \ref{eq:grad_pxy} frequently does when improper samples from the distribution are used. Notice that the output of the Softmax is always between $0$ and $1$ which is not necessarily true for Equation \ref{eq:energy_px} and Equation \ref{eq:energy_pxy} whose value can be arbitrarily large because its input $\x$ is continuous. This is even more likely to happen when SGLD is not providing competitive samples as their energy becomes very small. We hypothesize that this might be the source of the instability and we propose to find a different way to optimize JEM. %[We first propose to add MI in Section \ref{sec:mi} to achieve what we just described and then extend it to be used even without MI].

\subsection{Adding mutual information to the loss function}\label{sec:mi}
We propose to add mutual information (MI) of inputs $\mathbf{X}$ and labels $\text{Y}$ distributed according to $p_{\boldsymbol{\theta}}$ defined as
\begin{equation}\label{eq:mi}
\operatorname{I}(\mathbf{X};\text{Y}) = \KL(p_{\boldsymbol{\theta}}(\mathbf{x},y) \Vert p_{\boldsymbol{\theta}}(\mathbf{x})p_{\boldsymbol{\theta}}(y)) = \E_{(\mathbf{x},y) \sim p_{\boldsymbol{\theta}}(\mathbf{x},y)} \left[\log \left(\frac{p_{\boldsymbol{\theta}}(\mathbf{x},y)}{p_{\boldsymbol{\theta}}(\mathbf{x})p_{\boldsymbol{\theta}}(y)}\right) \right]
\end{equation}
to the original objective function ${\log p_{\boldsymbol{\theta}}(\mathbf{x},y)}$, therefore minimizing the loss function $\operatorname{L}$:
\begin{equation}\label{eq:true_objective}
-\operatorname{L} = -{\E_{(\mathbf{x},y) \sim p_{data}(\mathbf{x},y)}} \left[\log{p_{\boldsymbol{\theta}}(\mathbf{x},y)}\right] + \E_{(\mathbf{x},y) \sim p_{\boldsymbol{\theta}}(\mathbf{x},y)} \left[\log \left(\frac{p_{\boldsymbol{\theta}}(\mathbf{x},y)}{p_{\boldsymbol{\theta}}(\mathbf{x})p_{\boldsymbol{\theta}}(y)}\right) \right]
\end{equation}
Maximization of MI can also be interpreted as having a sharp posterior distribution $p_{\boldsymbol{\theta}}(y \mid \x )$ of each sample $\mathbf{x}$ while maximizing the entropy of marginal distribution $p(y)$ (Equation \ref{eq:mi_entr}). In other words, the global optimum is reached when each sample belongs only to a single class while all samples together are distributed uniformly with respect to class $y$.
\begin{equation}\label{eq:mi_entr}
\operatorname{I}(\mathbf{X};\text{Y}) = \E_{(\mathbf{x},y) \sim p_{\boldsymbol{\theta}}(\mathbf{x},y)} \left[\log \left(\frac{p_{\boldsymbol{\theta}}(y \mid \x )}{p_{\boldsymbol{\theta}}(y)}\right) \right] = \E_{(\mathbf{x},y) \sim p_{\boldsymbol{\theta}}(\mathbf{x},y)} \left[\log p_{\boldsymbol{\theta}}(y \mid \x ) \right] + \operatorname{H}(p_{\boldsymbol{\theta}}(y))
\end{equation}
Approximating the expectation over the model distribution $p_{\boldsymbol{\theta}}$ by the expectation over $p_{data}$ leading to a new loss function $\operatorname{L_a}$ (Equation \ref{eq:approx_objective}).
\begin{equation}\label{eq:approx_objective}
-\operatorname{L_a}={\E_{p_{data}}} \left[\log{p_{\boldsymbol{\theta}}(\mathbf{x},y)} + \log \left(\frac{p_{\boldsymbol{\theta}}(\mathbf{x},y)}{p_{\boldsymbol{\theta}}(\mathbf{x})p_{\boldsymbol{\theta}}(y)}\right) \right] = {\E_{p_{data}}} \left[\log{p_{\boldsymbol{\theta}}(y \mid \x )} + \log{p_{\boldsymbol{\theta}}(\mathbf{x} \mid y)} \right]
\end{equation}
%[Matthew said that taking these samples should be lower-bound of MI, in any case, https://arxiv.org/pdf/1905.06922.pdf - he told me to check this, it could be 2.3 but I am not sure if it is exactly the same thing]
The rationale behind the approximation of Equation \ref{eq:true_objective} by Equation \ref{eq:approx_objective} is that the the optimum of the first RHS term of Equation \ref{eq:true_objective} is reached when $p_{data}=p_{\boldsymbol{\theta}}$ and the term $\E_{p_{data}} \left[\log \left(\frac{p_{\boldsymbol{\theta}}(\mathbf{x},y)}{p_{\boldsymbol{\theta}}(\mathbf{x})p_{\boldsymbol{\theta}}(y)}\right) \right]$ share the same optimum when the dataset is balanced but does not motivate $p_{\thet}(y)$ to become an uniform distribution. Moreover, even before reaching the optimum, we want samples from ${p_{data}}$ to be the subset of samples from ${p_{\boldsymbol{\theta}}(y \mid \x )}$ and therefore we are performing almost correct updates for train data in order to maximize Equation \ref{eq:true_objective}.
We show how to get even closer to optimizing the true objective function (Equation \ref{eq:true_objective}) in Section \ref{sec:samples_xent}. 

\subsection{Maximizing $p_{\mathbf{\theta}}(\mathbf{x} \mid y)$}
%\subsection{Maximizing )}
Maximizing $p_{\boldsymbol{\theta}}( y \mid \x )$ can be simply achieved with a softmax function as described in Section \ref{sec:classification}. We are able to evaluate the exact posterior distribution $p_{\boldsymbol{\theta}}( y \mid \x )$ which is not influenced by $Z_{\theta}$. As mentioned before, the training is stable and we suggest that it is a consequence of the fact that the train data example is present both in the numerator and denominator and also the fact that we can evaluate the denominator for all values of $y$. We would like to have the same effect when maximizing $p_{\thet}(\x \mid y)$. Expressing $p_{\boldsymbol{\theta}}(\mathbf{x} \mid y)$ in Equation \ref{eq:px_giv_y} to resemble Equation \ref{eq:softmax}.
\begin{equation} \label{eq:px_giv_y}
p_{\boldsymbol{\theta}}(\mathbf{x} \mid y) = \frac{p_{\boldsymbol{\theta}}(\mathbf{x}, y)}{\int_{\mathbf{x}} p_{\boldsymbol{\theta}}(\mathbf{x}, y) d\mathbf{x}} = \frac{e^{f_{\boldsymbol{\theta}}(\mathbf{x})_y}}{\int_{\mathbf{x}} e^{f_{\boldsymbol{\theta}}(\mathbf{x})_y} d\mathbf{x}} 
\end{equation}
We could use the gradient\footnote{Detailed derivation of Equation \ref{eq:px_giv_y_grad} can be found in Appendix (Equation \ref{eq:px_giv_y_grad_full}).} of $\log p_{\boldsymbol{\theta}}(\mathbf{x} \mid y)$ to maximize $p_{\thet}(\x \mid y)$:
\begin{equation} \label{eq:px_giv_y_grad}
\nabla_{\boldsymbol{\theta}} \log p_{\boldsymbol{\theta}}(\mathbf{x} \mid y) = \nabla_{\boldsymbol{\theta}} f_{\boldsymbol{\theta}}(\mathbf{x})_y - \E_{\mathbf{x} \sim p_{\boldsymbol{\theta}}(\mathbf{x} \mid y)} \left[\nabla_{\boldsymbol{\theta}} f_{\boldsymbol{\theta}}(\mathbf{x})_y \right] \approx \nabla_{\boldsymbol{\theta}} f_{\boldsymbol{\theta}}(\mathbf{x})_y - \frac{1}{N} \sum_i \nabla_{\boldsymbol{\theta}} f_{\boldsymbol{\theta}}(\mathbf{x^i})_y
\end{equation}
Although this resembles Equation \ref{eq:px_giv_y_grad}, using this equation to compute gradients will unfortunately lead to the same instability issues. We suggest an alternative way, by first realizing that in order to calculate the intractable part of Equation \ref{eq:px_giv_y_grad}, we can alternatively use a different distribution $q(\mathbf{x})$ and eventually approximate it using importance sampling. In a special case when the distribution $q{(\x)}$ is (continuous) uniform $q_u(\mathbf{x})$, there is a constant $\operatorname{k}$ such that
\begin{equation} \label{eq:px_giv_y_grad_by_q}
\E_{\mathbf{x} \sim p_{\boldsymbol{\theta}}(\mathbf{x} \mid y)} \left[\nabla_{\boldsymbol{\theta}} f_{\boldsymbol{\theta}}(\mathbf{x})_y \right] = \E_{\mathbf{x} \sim q(\mathbf{x})} \left[ \frac{p_{\boldsymbol{\theta}}(\mathbf{x} \mid y)}{q(\mathbf{x})} \nabla_{\boldsymbol{\theta}} f_{\boldsymbol{\theta}}(\mathbf{x})_y \right] = \operatorname{k}\: \E_{\mathbf{x} \sim q_u(\mathbf{x})} \left[ p_{\boldsymbol{\theta}}(\mathbf{x} \mid y) \nabla_{\boldsymbol{\theta}} f_{\boldsymbol{\theta}}(\mathbf{x})_y \right] 
\end{equation}
We hypothesize that in the worst case (complete failure), SGLD draws samples from $q_u(\mathbf{x})$ instead of the desired distribution (in this case $p_{\thet}(\x \mid y)$, alternatively $p_{\thet}(\x)$ or $p_{\thet}(\x, y)$). If that happens, we should weight samples by $p_{\thet}(\x \mid y)$. Approximating expectation of $q_u(\mathbf{x})$ by samples for which we cannot evaluate the exact likelihood of $p_{\thet}(\x \mid y)$ in our model, we are forced to use self-normalized variant of importance sampling \cite{bishop2006pattern}. Expressing Equation \ref{eq:px_giv_y_grad_by_q} using self-normalized importance sampling, we get
\begin{equation} \label{eq:px_giv_y_grad_exp_alter}
%\operatorname{k}\: \E_{\mathbf{x} \sim q_u(\mathbf{x})} \left[ p_{\boldsymbol{\theta}}(\mathbf{x} \mid y) \nabla_{\boldsymbol{\theta}} f_{\boldsymbol{\theta}}(\mathbf{x})_y \right] = 
\operatorname{k} \E_{\mathbf{x} \sim q_u(\mathbf{x})} \left[ \frac{p_{\thet}(\x,y)}{p_{\thet}(y)} \nabla_{\boldsymbol{\theta}} f_{\boldsymbol{\theta}}(\mathbf{x})_y \right] \approx \sum_i^N  \frac{\operatorname{k} \frac{p_{\thet}(\mathbf{x^i},y)}{p_{\thet}(y)}}{\operatorname{k} \sum_j^N \frac{p_{\thet}(\mathbf{x^j},y)}{p_{\thet}(y)}} \nabla_{\thet} f_{\thet}(\mathbf{x^i})_y  = \sum_i^N \frac{e^{f_{\boldsymbol{\theta}}(\mathbf{x^i})_y}}{\sum_j^N e^{f_{\boldsymbol{\theta}}(\mathbf{x^j})_y}} \nabla_{\thet} f_{\thet}(\mathbf{x^i})_y
\end{equation}
Following this strategy when SGLD fails for $\textbf{all}$ samples, it can still be problematic as the denominator of Equation \ref{eq:px_giv_y} might be much smaller than its numerator. Since the correct denominator of Equation \ref{eq:px_giv_y_grad} should be much larger than estimated by our (incorrect) samples, we are scaling this gradient compared to the situation when at least one of the samples has comparable energy to train data (this is the consequence of using a self-normalized variant of importance sampling). We can overcome this issue of not estimating the denominator correctly by including the real (training) data point for which we are computing the gradient as one of the negative samples which works as an anchor\footnote{Or as a wall in the analogy introduced in Section \ref{sec:contrib}}. The trick of including a positive sample into negative ones $\textbf{cannot}$ be used when using an approximation from Equation \ref{eq:px_giv_y_grad} because we would ignore the gradient of training data by simply subtracting exactly the same gradient. This gives us a stable way to maximize Equation \ref{eq:px_giv_y_grad} in a case when SGLD produces improper samples. Since the negative energy of generated samples will be much smaller than the real ones, we effectively ignore these samples because the computation of the gradient from Equation \ref{eq:px_giv_y_grad} becomes almost $\nabla_{\boldsymbol{\theta}}f_{\boldsymbol{\theta}}(\mathbf{x})_y - \nabla_{\boldsymbol{\theta}} f_{\boldsymbol{\theta}}(\mathbf{x})_y$, i.e. zero. 

Notice that if EBM tries to reach a likelihood close to 0 for most of the input points $\x$ and we assume that they are effectively exactly 0, we don't need to estimate the gradient based on these. More than that, if we assume that the rest of the input points lie on a low-dimensional manifold either having the same likelihood or the same probability of being sampled by SGLD, the suggested update follows the gradient. We do not think that these assumptions hold and it leads us to an investigation of what we will optimize following suggested updates shall the SGLD provide proper samples. 

%In JEM, authors use samples from approximated SGLD as samples from the desired distributions (here $p_{\boldsymbol{\theta}}(\mathbf{x} \mid y)$). An updated version of SGLD can sometimes produce low-quality samples with low negative energy. In the worst case, samples can come from $q_u(\mathbf{x})$. If that is the case, it is correct to scale the gradients by $p_{\boldsymbol{\theta}}(\mathbf{x} \mid y)$. Due to unknown $Z_{\boldsymbol{\theta}}$ it would be necessary to apply a self-normalized variant of importance sampling. [I have it later, that should be rearranged somehow or said that it will be shown later]

Before we do that, let us first show what we optimize by using Equation \ref{eq:px_giv_y_grad} in the case when SGLD generate improper samples distributed according to $q_u(\mathbf{x})$. As seen from Equation \ref{eq:px_giv_y_grad_by_q}, we are effectively scaling their gradient proportionally to $\frac{1}{p_{\boldsymbol{\theta}}(\mathbf{x} \mid y)}$, which will lead to the estimator with undesirably high bias and we suspect it to be the main reason of instability\footnote{One of the recommended tricks to stabilize the training is to ignore samples that have very low likelihood and try to enforce generated samples to have a comparable likelihood to training data e.g. by directly adding the minimization of the difference to the objective function.}. 

Notice that when maximizing Equation \ref{eq:softmax}, the value itself cannot be larger than 1 (numerator is not larger than denominator). Practically, this holds because we enumerate over all possibilities of $y$ and the one in the numerator is always part of the denominator. When we try to approximate intractable integral in \ref{eq:px_giv_y}, the same is not true. Therefore, we suggest including the data point from the numerator in the denominator. Moreover, we recommend taking advantage of the fact that other data points in the mini-batch are also samples from $p_{data}$ and the goal of the training is that they are actual samples from $p_{\boldsymbol{\theta}}$. We have observed that including all samples from mini-batch increases the performance\footnote{We even include  samples that belong to a different class according to $p_{data}$ as the desired value under $p_{\thet}$ is anyway $0$.} as shown in Appendix \ref{sec:exp_images} and is also faster to compute and easier to implement.

Let us now come back and explain the effect of plugging Equation \ref{eq:px_giv_y_grad_exp_alter} into \ref{eq:px_giv_y_grad} when SGLD behaves as expected and provides samples from desired distribution (in our case $p_{\thet}(\x \mid y)$). In this case, each $\x$ in the denominator of $\frac{e^{f_{\boldsymbol{\theta}}(\mathbf{x})_y}}{\int_{\mathbf{x}} e^{f_{\boldsymbol{\theta}}(\mathbf{x})_y} d\mathbf{x}}$ is multiplied by $p_{\boldsymbol{\theta}}(\mathbf{x} \mid y)$ leading to $\frac{e^{f_{\boldsymbol{\theta}}(\mathbf{x})_y}}{\int_{\mathbf{x}} p_{\boldsymbol{\theta}}(\mathbf{x} \mid y) e^{f_{\boldsymbol{\theta}}(\mathbf{x})_y} d\mathbf{x}}$. Original training of the JEM has the analogy that we push up (trying to increase the likelihood) on training data (positive examples) while we push down (trying to increase the likelihood) on generated (negative) examples\footnote{Or vice versa when we talk about energy instead of likelihood.}. In this analogy, we now push down on the negative examples proportionally to their likelihood more\footnote{Notice, that originally we push on the same on all the examples, but these examples are selected based on their likelihood.}. Better intuition can be gained by realizing that the contribution coming from negative samples (in the limit, when an infinite amount of them are used) will not influence $p_{\thet}(\x^1) \ge p_{\thet}(\x^2)$ for two input data points $\x^1$ and $\x^2$. This means that if the condition holds before the update, it will hold after the update. Our proposed update rule does not have the same property, but in practice, when taking a limited amount of samples, this property is not guaranteed even for the original method. Furthermore, realize that we are updating the parameters $\thet$ in an iterative manner, meaning that compared to the update suggested by the gradient, we push down more on likely $\x$, which results in them becoming less likely. As a consequence, we are going to push proportionally less on them in the next iteration if they are chosen as negative samples. Equation \ref{eq:change_in_update_rule} visualizes the resulting change in the update compared to previous technique. Notice that implementation-wise we just first generate SGLD samples, include them in mini-batch and then use the softmax function over the mini-batch. 
\begin{equation}\label{eq:change_in_update_rule}
\nabla_{\boldsymbol{\theta}} f_{\boldsymbol{\theta}}(\mathbf{x})_y - \frac{1}{N} \sum_i \nabla_{\boldsymbol{\theta}} f_{\boldsymbol{\theta}}(\mathbf{x^i})_y \; \rightarrow \; \nabla_{\boldsymbol{\theta}} f_{\boldsymbol{\theta}}(\mathbf{x})_y - \sum_i^N \frac{e^{f_{\boldsymbol{\theta}}(\mathbf{x^i})_y}}{\sum_j^N e^{f_{\boldsymbol{\theta}}(\mathbf{x^j})_y}} \nabla_{\boldsymbol{\theta}} f_{\boldsymbol{\theta}}(\mathbf{x^i})_y  
\end{equation}
The goal of the training is to maximize the expectation over the training data $\E_{p_{data}} \left[ \frac{e^{f_{\boldsymbol{\theta}}(\mathbf{x})_y}}{\int_{\mathbf{x}} e^{f_{\boldsymbol{\theta}}(\mathbf{x})_y} d\mathbf{x}} \right]$. The global (over-trained) optimum is reached when the energy of each sample from the dataset is exactly the same (assuming each data point is at maximum once in the dataset) and $0$ everywhere else. Assuming that the model $f_{\thet}(\x,y)$ is powerful enough to reach this optimum, this optimum is the same even when following updates of proposed method in Equation \ref{eq:px_giv_y_grad_exp_alter}. 
\begin{equation} \label{eq:p_x_optimum}
\argmax_{\boldsymbol{\theta}} \frac{e^{f_{\boldsymbol{\theta}}(\mathbf{x})_y}}{\int_{\mathbf{x}} e^{f_{\boldsymbol{\theta}}(\mathbf{x})_y} d\mathbf{x}} = \argmax_{\boldsymbol{\theta}} \frac{e^{f_{\boldsymbol{\theta}}(\mathbf{x})_y}}{\int_{\mathbf{x}} p_{\boldsymbol{\theta}}(\mathbf{x} \mid y) e^{f_{\boldsymbol{\theta}}(\mathbf{x})_y} d\mathbf{x}}
\end{equation} 
Unfortunately, when a particular train data point occurs multiple\footnote{If each data point occurs exactly the same number of times -- such as when iterating multiple times over the same dataset -- Equation \ref{eq:p_x_optimum} still holds.} times in the dataset, this will not hold as we will push down on that specific data point more (compared to those less frequent). As this case might not be very important for the real-case scenarios, the real goal of training is not to reach the global over-trained optimum, we are approximating expectations by sampling and these samples will not be exact. 

\section{Discussion}
We can intuitively see this optimization as we are playing a game of showing multiple input samples to our model and asking it to ``bet unnormalized log-likelihood'', which should be ``balanced'' among that group of inputs. Afterward, ``bets'' that were not put on the correct class or were put on negative samples, are lost. The model is trained to regain back an as large portion of ``bets'' as possible. In this analogy, our approach forces the model to pay more attention to difficult examples. We show later that this method indeed stabilizes the training and, in fact, increases the accuracy of the trained model. We further investigate the reason for the increase in accuracy. To distinguish the model trained by the proposed approach, we refer to it as Stable Training JEM (ST-JEM).

Likewise, an analogy can be made for training. Original training of JEM can be compared to two people (positive and negative samples) lifting a large ball by pushing on it from the opposite sides (the ball can move higher only when both are pushing). When one (negative samples) is not providing comparable force, the force balance is disturbed and the ball falls down (training diverges). We propose that both people should push from the same side using a wall (softmax cannot be higher than 1) on the other side so when the weaker (negative samples) cannot push hard enough, the stronger simply keeps the ball in the same position. It provides time for the weaker to regain its power (details are discussed in Appendix \ref{sec:sgld_hyper_parameters}). Moreover, our method could be extended to use any (negative) samples for which low log-likelihood is desired, such as those from out-of-domain datasets or those provided by VERA.

We have performed many experiments to help us better understand the system. Due to the space limitation, we share our insight in the Appendix. As we are not required to use any amount of SGLD samples, we investigate the effect of not having any in Appendix \ref{sec:res_jem}
Experimental part of this work is in Appendix \ref{sec:exp_speech} and Appendix \ref{sec:exp_images}. The rest of the Appendix describes additional variants and implementation details.

\section{Conclusion}
We introduced an alternative way of training JEM, called ST-JEM. Unlike the previous training of JEM, our training does not diverge. This is done by making it robust to the effect of improper samples that are the consequence of approximation of the SGLD procedure. Following the proposed training, we obtained systems on two modalities -- images and speech. For images, we show that the introduced way of training not only stabilizes the training but also increases the classification accuracy. By investigation, we localized the source of increase of the accuracy, which is not a consequence of the generative part of JEM, but the way of training. Our method allows for any number of SGLD samples per mini-batch and we show that the case of not using any (RES-JEM), leads to the same or even slightly better classification performance than the one reached by ST-JEM. We show that RES-JEM can still be interpreted as JEM that is defined only over the domain of training data points. This assumption makes the model generative only on the theoretical level, as we can neither sample from it, nor evaluate the likelihood outside of the defined domain. In practice, we still use it outside of the defined domain as a classifier that reaches superior accuracy over the discriminative model on CIFAR-10, CIFAR-100, and also when applied to speech. The effectiveness of this approach is most evident for low-resource or more difficult problems. We also demonstrate that, unlike JEM, we are able to increase accuracy by incorporating unlabeled data. Last but not least, we provide a discussion in Appendix \ref{sec:speech_jem_future} and suggest possible future usage of ST-JEM, which suggests slightly shifting the currently popular approach of training a system on the task and then performing inference by simply forwarding the input through the model with fixed computation time to more human-like. We suggest that many problems can be reformulated as a search problem using trained ST-JEM and we propose to have ST-JEM that will be able to model multiple joint distributions at the same time, which further constrain the search problem and should lead to improved performance by introducing a bias such that the segments that are not speech should not contribute to the process of speaker identification.

\bibliography{iclr2023_conference}
\bibliographystyle{iclr2023_conference}

\appendix

\section{Variations of ST-JEM}

\subsection{RES-JEM}\label{sec:res_jem}
If we don't use any SGLD samples, we end up with a method called RES-JEM. Since we approximate negative samples by mini-batch examples, we can perform the same updates even in the case of including $0$ SGLD samples. As this approach seems vague, we explain that it is well-defined and we demonstrate that it actually performs better than a discriminative model maximizing cross-entropy. RES-JEM can be understood as a JEM, whose domain is only a subset of $\mathbf{X}$, where all training examples lie. We are sampling from the dataset that does not have duplicates and based on the definition of the subset of $\mathbf{X}$, which is exactly $q_u$ because we assume that that $p_{\thet}(\x)=0$ elsewhere., we have a special case  Because we use \ref{eq:approx_objective}, we are using expectation over training data. As explained in \ref{eq:px_giv_y_grad_by_q} so we have a In this definition,  Because of that if we use all training examples in a single batch, we directly optimize \ref{eq:true_objective} as our negative samples are now other training examples. The resulting approach is similar to \cite{liu2020hybrid}, where they used stored values of other logits from the dataset to re-weight the gradient

%whole batch
%sgld hpyer parameters

\subsection{Unlabeled data / Clustering / Unsupervised and Semi-Supervised Learning}
We can try to extend Eq. \ref{eq:approx_objective} in the case when we are missing labels. A natural extension is to apply an expectation over $y$ for samples for which we do not have ground truth labels.

\begin{equation}\label{eq:objective_unlabelad}
\E_{\mathbf{x} \sim p_{data}(\mathbf{x})} \left[ \overbrace{{\sum_y p_{\boldsymbol{\theta}}(y \mid \x )}}^{\text{ replaces } \; p_{data}(y \mid \x )} \left[ \log{p_{\boldsymbol{\theta}}(y \mid \x )} + \log{p_{\boldsymbol{\theta}}(\mathbf{x} \mid y)} \right] \right]
\end{equation}
Intuitively, this objective is maximized when the data are grouped into clusters, the number of clusters corresponds to the number of classes, and when sampling from the $p(x)$, the distribution of $y$ should be uniform. Notice that the same cannot be applied to Equation \ref{eq:discriminative} as the naive solution is to assign/collapse everything to a single class. 
Formed clusters can in general have any meaning and we suggest combining both Equation \ref{eq:approx_objective} and Equation \ref{eq:objective_unlabelad} in terms of semi-supervised learning to constrain the meaning of the clusters (in speech, clustering could be based on e.g. phonemes, speakers, frequencies, energies or just noise).

\subsection{Minimizing cross-entropy of conditionally generated samples}\label{sec:samples_xent}
By optimizing Eq. \ref{eq:approx_objective} instead of Eq. \ref{eq:true_objective}, we no longer enforce samples from $p_{\boldsymbol{\theta}}$ to belong to one class. We suggest to replace $p_{\boldsymbol{\theta}}(y)$ by $p_{data}(y)$ which leads to just maximizing posterior distribution $p_{\boldsymbol{\theta}}(\mathbf{x} \mid y)$ of class $y$ that the generation was conditioned on, because $H(y)$ becomes a constant. Note that it can be simply interpreted as minimizing cross-entropy but also can be seen as encoder-decoder ($y \rightarrow \x \rightarrow y$). Adding the following term to the loss function causes images sampled from $p_{\boldsymbol{\theta}}(x)$ (not conditioned on $y$) to more likely resemble objects of some particular class [reference to image comparison].
\begin{equation}\label{eq:samples_xent}
-\operatorname{L_{s}} = \E_{y \sim p_{data}(y)} \E_{\mathbf{x} \sim p_{\boldsymbol{\theta}}(\mathbf{x} \mid y)} \left[\log \left(\frac{p_{\boldsymbol{\theta}}(\mathbf{x},y)}{p_{\boldsymbol{\theta}}(\mathbf{x})p_{\boldsymbol{\theta}}(y)}\right) \right] = \E_{y \sim p_{data}(y)} \E_{(\mathbf{x}) \sim p_{\boldsymbol{\theta}}(\mathbf{x} \mid y)} \left[\log p_{\boldsymbol{\theta}}(y \mid \x ) \right]
\end{equation}
Adding the following objective can rarely cause the model to stop producing reasonable samples and we suspect that it happens when SGLD doesn't produce competitive samples yet we still minimize cross-entropy even for them. For that reason, we decided to run the experiments without this loss and only. %We noticed that when optimized with this additional loss, the quality of unconditionally generated samples increased its quality. %[neg log sum exp?]
%[Better to enforce to samples as it can end up in weird point when samples are low-quality but the posterior is high which is what the EBM likes - it is getting low-energy samples which has peaky posterior and it gets rewarded for both...]
%[Or maybe do the similar thing]

\subsection{Quality of SGLD samples and adaptive change of SGLD hyper-parameters during training}\label{sec:sgld_hyper_parameters}
"Optimal" hyper-parameters change during the training and we can afford to slightly adapt towards this change because we can afford some exploration (not just exploitation) because if we start producing bad samples, it doesn't affect the training... for that period of training we just maximize cross-entropy. In practice, we notice that there might be stages when we are not able to produce reasonable samples and then it gets better again - there is one huge problem - if we focus on getting the best log-likelihood, in the buffer are some samples that are already pretty good and some of them are bad. The hyper-parameters for the bad ones and good ones might be dramatically different in order to increase their negative energy. %[potential future, use RL or another approach to predict better hyper-parameters to use - for example, gan-style]

%\subsection{How many negative samples and steps do we need - speeding things up}
%We can take as much as we want if we want nice samples, and only few to get some samples

\section{Image Experiments}\label{sec:exp_images}
We train our systems ST-JEM and RES-JEM on CIFAR-10 and CIFAR-100 following the same procedure described in JEM. As we did not obtain exactly the same system and we were not able to train the JEM model, we report the results from the paper together with our results. As reported in Table \ref{tab:cifars}, even though our baseline is worse, our ST-JEM and RES-JEM work better than the reported baseline from JEM. We see a small drop in the performance of the ST-JEM compared to RES-JEM. Following the setup from JEM in low-resource settings, we use only 4000 labeled examples for CIFAR-10 and CIFAR-10 and report the largest boost in the accuracy compared to previous systems in Table \ref{tab:cifars_4k}. We also investigate effect of other hyper-parameters in Table \ref{tab:cifars_oth}.

%[Rearrange - I want to show
%a) BS 64 -> BS 256 doesn't have influence -> it has influence but the same for baseline and MI
%b) no positive samples is worse than when included
%c) More samples better quality of samples but worse accuracy/calibration
%d) When using much more steps, we can reach the same accuracy, it seems that the accuracy drop is because we are not using so good samples - getting them can increase accuracy
%e) XENT over p(y|x) samples causes to produce better samples (will have sharper posterior) when sampling from p(x)
%f) GP - gradient penalty restricts the model, might be easier to sample from SGLD but might also hurt the performance - we see how it hurts baseline, mi, jem; It seems that even applying this regularization, JEM can reach similar accuracy. I am not sure if it is a fair comparison because relatively, it might be larger weight for baseline, then MI and then JEM. GP seemed to be helpful for speech experiments so we use it there.
%g) 

\begin{table}[H]
\centering
\renewcommand{\arraystretch}{1.2}
\caption{Classification accuracies [\%] and ECE [\%] values for the CIFAR-10 and CIFAR-100 data sets are shown. Apart from the baseline and JEM results reported in \cite{grathwohl2019your}, our results with RES-JEM and the DEFAULT ST-JEM models which uses 8 SGLD samples and 20 SGLD steps are also shown.}
\begin{tabular}{@{}lcccc@{}}
\cmidrule(l){2-5}
                      & \multicolumn{4}{c}{\textbf{Data sets}}                                                                                                        \\ \cmidrule(l){2-5} 
                      & \textbf{CIFAR-10} & \multicolumn{1}{c|}{\textbf{CIFAR-100}} & \multicolumn{1}{l}{\textbf{CIFAR-10}} & \multicolumn{1}{l}{\textbf{CIFAR-100}} \\ \cmidrule(l){2-5} 
\textbf{Model}        & \multicolumn{2}{c|}{\textbf{Accuracy \%}}                   & \multicolumn{2}{c}{\textbf{ECE \%}}                                            \\ \midrule
Baseline (reported)   & 93.6              & \multicolumn{1}{c|}{74.2}               & 2.7                                   & 22.3                                   \\
Baseline (ours)       & 93.5              & \multicolumn{1}{c|}{72.0}               & 5.4                                   & 22.6                                   \\ \midrule
JEM (reported)        & 92.9              & \multicolumn{1}{c|}{72.2}               & 2.9                                   & 4.9                                    \\
RES-JEM (ours)         & 94.2              & \multicolumn{1}{c|}{76.7}               & 4.4                                   & 7.6                                    \\
DEFAULT ST-JEM (ours) & 93.9              & \multicolumn{1}{c|}{75.7}               & 4.6                                   & 11.6                                   \\ \bottomrule
\end{tabular}
\label{tab:cifars}
\end{table}

\begin{table}[H]
\centering
\renewcommand{\arraystretch}{1.4}
\caption{The table shows the classification accuracy [\%] and ECE [\%] values for the CIFAR-10 and CIFAR-100 data sets with 4k labels.}
\begin{tabular}{lcccc}
\cline{2-5}
               & \multicolumn{4}{c}{\textbf{Datasets (4k labels)}}                                                                                            \\ \cline{2-5} 
               & \textbf{CIFAR 10} & \multicolumn{1}{c|}{\textbf{CIFAR 100}} & \multicolumn{1}{l}{\textbf{CIFAR 10}} & \multicolumn{1}{l}{\textbf{CIFAR 100}} \\ \cline{2-5} 
\textbf{Model} & \multicolumn{2}{c|}{\textbf{Accuracy \%}}                   & \multicolumn{2}{c}{\textbf{ECE \%}}                                            \\ \hline
Baseline (reported)   &     78.0          & \multicolumn{1}{c|}{}               &           18.2                         &                                   
\\ 
Baseline       & 77.4              & \multicolumn{1}{c|}{33.9}               & 18.3                                  & 51.4                                   
 \\\hline
JEM (reported)        &      74.9         & \multicolumn{1}{c|}{}               & 13.7                                   &                                     \\
RES-JEM         & 81.1              & \multicolumn{1}{c|}{38.3}               & 13.8                                  & 23.0                                   \\
DEFAULT ST-JEM & 79.4              & \multicolumn{1}{c|}{35.4}               & 14.3                                  & 19.2                                   \\ \hline
&   \multicolumn{4}{c}{\textbf{Including unlabeled data}}   
\\ \hline
RES-JEM & 82.53              & \multicolumn{1}{c|}{}               & 15.47                                  &                                   
\\ \hline
\end{tabular}
\label{tab:cifars_4k}
\end{table}

\begin{table}[H]
\centering
\caption{Results from variations of the DEFAULT ST-JEM model are shown (see Section \ref{sec:exp_images}). Although for the default ST-JEM configuration, the number of SGLD steps is fixed, further experimentation is done with a variable number of SGLD steps, linearly adding +1 step every 1/10 epochs. Additional experimental results with gradient penalty and incorporation of cross-entropy loss for better image generation are also shown.}
\renewcommand{\arraystretch}{1.5}
\begin{tabular}{lcc}
\hline
\textbf{ST-JEM variations on CIFAR-100 data set}                                                                              & \textbf{Accuracy \%} & \textbf{ECE \%}      \\ \hline
DEFAULT - 8 SGLD samples + fixed 20 SGLD steps                                                                                      & 75.7                 & 11.6                 \\
\textbf{changing the number of SGLD samples}                                                                                  &                      &                      \\ \hline
2 SGLD samples + gradually increasing SGLD steps from 20 to 40                                                                & 76.0                 & 10.9                 \\
8 SGLD samples + gradually increasing SGLD steps from 20 to 40                                                                & 75.6                 & 13.3                 \\
32 SGLD samples + gradually increasing SGLD steps from 20 to 40                                                               & 74.3                 & 14.4                 \\
\textbf{changing the range of SGLD steps}                                                                                    &                      &                      \\ \hline
8 SGLD samples + gradually increasing SGLD steps from 20 to 220                                                               & 76.5                 & 13.5                 \\
\textbf{adding gradient penalty}                                                                                              & \multicolumn{1}{l}{} & \multicolumn{1}{l}{} \\ \hline
\begin{tabular}[c]{@{}l@{}}8 SGLD samples + gradually increasing SGLD steps from 20 to 220 \\ + gradient penalty\end{tabular} & 76.1                 & 13.9                 \\
\textbf{adding cross entropy}                                                                                                 & \multicolumn{1}{l}{} & \multicolumn{1}{l}{} \\ \hline
DEFAULT + XENT                                                                                                                & 75.2                 & 13.6                 \\ \hline
\end{tabular}
\label{tab:cifars_oth}
\end{table}

\section{Speech Experiments}\label{sec:exp_speech}
We trained ST-JEM to model the joint distribution of $21$ consecutive $80$ dimensional Mel-filterbank $\x$ and phoneme $y$ corresponding to the middle frame. We demonstrate the benefits of ST-JEM when combining models trained on different datasets in Section \ref{sec:sp_comb}. We further demonstrate that the system is capable of producing reasonable speech by original or modified SGLD procedure. The speech is produced by creating a sequence of $200$ frames. For each $21$ consecutive frames, we are able to predict joint distribution $p_{\thet}(x,y)$ and we maximize the likelihood of the overall sequence by pretending that they are independent as they are already dependent through heavy overlap in $\x$. Conditioned on a sequence of 180 randomly extracted labels from the test set. We are, somewhat surprisingly, able to produce understandable speech by first obtaining an approximate spectrogram through pseudoinverse of the transformation from spectrogram to mel-filterbank and then reconstructing the phase of that spectrum by using only Grififn-Lim algorithm. We are attaching samples to the submission. When the review process is over, we will publish the code together with all samples. By slight modification of SGLD procedure to make it more greedy during the inference, we are able to denoise heavily corrupted speech, samples are also attached to the submission. Preliminary evaluation on a few examples showed that while ST-JEM is able to reach about $75\%$ of accuracy, heavily corrupting a small portion of input (such that every 21 consecutive frames are affected, results in accuracy around $1-20\%$. By applying a similar procedure to SGLD, we were typically able to reach an accuracy of about $35-50\%$. Further and more systematic evaluation is needed to confirm that ST-JEM can work as a denoiser in a realistic setting, but our preliminary results suggest so. 

\subsection{Model Combination} \label{sec:sp_comb}
Traditional machines learning systems show unsatisfactory generalization to unknown data domains. The widely prevalent solutions to this unavoidable problem are a variety of data-augmentation methods or, in other cases, just the accumulation of training data over as many test conditions as possible for model training. However, an often overlooked solution is that via model combination \cite{ensemble,kuncheva2014combining,sadhu2020continual,sustek22_interspeech}. 

Given two classifiers trained on different data domains, under an unknown test condition the aim is to obtain a \textit{weighted combination} of the posterior distribution from the two models that lead to better classification accuracy. That said, the main challenge in this approach lies in finding the best combination strategy.  In the following section, we describe our combination strategy with ST-JEM-based speech recognition systems. 

\subsubsection{Combination of Automatic Speech Recognition (ASR) systems}

Hybrid automatic speech recognition systems require conditional likelihood values $p(\mathbf{x}|y)$ for every feature vector $\mathbf{x}$ computed at a desired temporal sampling rate over a considerable duration of speech for a large ($\approx 3000$) number  of tri-phonetic states $y$ of a Hidden Markov Model (HMM). During inference, these likelihood values are then passed onto a decoding graph to find the best path evaluated by likelihoods and constrained by context, lexicon, and grammar to obtain text \cite{povey2011kaldi}.

Consider two ST-JEMs trained on two different data sets indexed by 1 and 2. For our experiments, the two data sets used are Wall Street Journal (WSJ) and REVERB which comprise clean read speech and simulated reverberated speech respectively. The input features $\mathbf{x} \in \mathbb{R}^{80 \times 9}$ are obtained by concatenating $80$ dimensional Mel-filterbank features over $9$ contiguous frames sampled at $100$ Hz in time with $3376$ triphone states i.e., $y \in \{0,1,2, \dots 3375\}$. Assigning $\boldsymbol{\theta_1}$ and $\boldsymbol{\theta_2}$ to be the learned parameters from the first and second data set respectively and $\boldsymbol{\theta_{comb}}$ to be the combined parameter set, we propose the following combination strategy.
\begin{eqnarray}
\label{eq:combination_strategy}
    p_{\boldsymbol{\theta_{comb}}}(\x,y)&=& \frac{p_{\boldsymbol{\theta_1}}(\x,y)+p_{\boldsymbol{\theta_2}}(\x,y)}{2} \\ \nonumber 
    &=& \frac{e^{f_{\boldsymbol{\theta_1}}(\mathbf{x})_y}}{2 Z(\boldsymbol{\theta_1})}+\frac{e^{f_{\boldsymbol{\theta_2}}(\mathbf{x})_y}}{2 Z(\boldsymbol{\theta_2})}
\end{eqnarray}

To understand the rational behind combining joint distributions, observe that $p_{\boldsymbol{\theta_{comb}}}(\mathbf{x},y)=\frac{p_{\boldsymbol{\theta_1}}(\mathbf{x})p_{\boldsymbol{\theta_1}}(y | \mathbf{x})+p_{\boldsymbol{\theta_2}}(\mathbf{x})p_{\boldsymbol{\theta_2}}(y|\mathbf{x})}{2}$. Therefore $p_{\boldsymbol{\theta_{comb}}}(\mathbf{x},y)$ automatically combines the posterior distributions from each model \textit{weighted} by the likelihood of a given feature vector $\mathbf{x}$ from each model. For an unknown test feature vector $\mathbf{x}_{test}$, the relative value of $p_{\boldsymbol{\theta_1}}(\mathbf{x}_{test})$ vs $p_{\boldsymbol{\theta_2}}(\mathbf{x}_{test})$ represents how well the two ST-JEMs recognize $\mathbf{x}_{test}$ to match with their individual training conditions - a higher likelihood indicating a better match.

In our experiments, we observed that the partition functions from two different JEM-STs have a very similar range of values obtained over several SGLD samples which leads us to safely assume $Z(\boldsymbol{\theta_1}) \approx Z(\boldsymbol{\theta_2})$. This simplification is further motivated by the fact that preserving Equation \ref{eq:combination_strategy} as is for the combination strategy makes no consequential change in the final ASR performance. 

For some constant value $C$, assuming $Z(\boldsymbol{\theta_1})= Z(\boldsymbol{\theta_2})=C$, we get 

\begin{eqnarray}
\label{eq:combination_strategy2}
    \log p_{\boldsymbol{\theta_{comb}}}(\mathbf{x},y)&=& \log ( e^{f_{\boldsymbol{\theta_1}}(\mathbf{x})_y}+e^{f_{\boldsymbol{\theta_2}}(\mathbf{x})_y} ) - \log 2 C
\end{eqnarray}

The constant $C$ being simply a scaling factor, and given the prior probability distribution $p(y)$, the conditional log-likelihoods required by the decoding graph can be obtained as follows

\begin{eqnarray}
\label{eq:combination_strategy3}
    \log p_{\boldsymbol{\theta_{comb}}}(\mathbf{x} | y)&\equiv& \log ( e^{f_{\boldsymbol{\theta_1}}(\mathbf{x})_y}+e^{f_{\boldsymbol{\theta_2}}(\mathbf{x})_y} ) - \log p(y)\text{.}
\end{eqnarray}
Note that our combination strategy comes down to computing the logsumexp of the logits across models and can be easily generalized to more than two ST-JEMs as in Equation \ref{eq:combination_strategy3}.

\begin{eqnarray}
\label{eq:combination_strategy4}
    \log p_{\boldsymbol{\theta_{comb}}}(\mathbf{x} | y)&\equiv& \log \sum_i \exp e^{f_{\boldsymbol{\theta_i}}(\mathbf{x})_y} - \log p(y)
\end{eqnarray}
Table \ref{tab:asr_performance} shows a comparison of Word Error Rates (WER \%) of baseline and ST-JEM model combination. The baseline combination follows the same principle as Equation \ref{eq:combination_strategy3} and \ref{eq:combination_strategy4} where the ST-JEM logits are replaced by the logits from a standard classifier with the same architecture as the ST-JEM model.

\begin{figure}[h]
  \centering
  \includegraphics[width=0.95\textwidth]{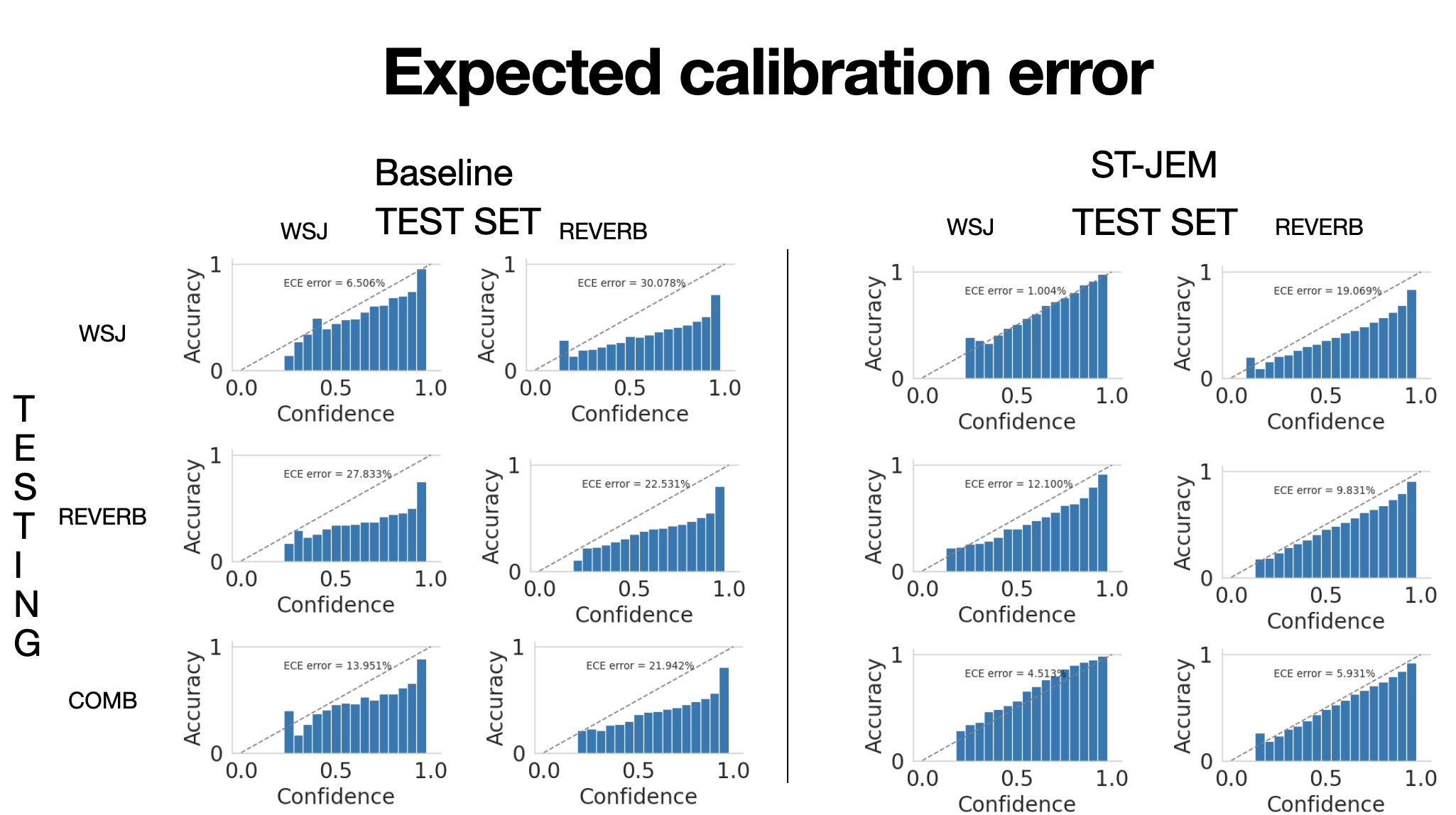}
  %\includegraphics[width=0.45\textwidth]{figs/inp_wer_ll_correlation.pdf}
  %\vspace{-1.5em}
  \caption{Comparison of ECE between Baseline system and proposed ST-JEM evaluated on WSJ and REVERB datasets. The last row corresponds to the described combination of the systems.}
  \label{fig:ece_speech_comb}.
  %\vspace{-1em}
\end{figure}

\begin{table}[H]
\caption{ASR Word Error Rate [\%] results are shown for individual ST-JEM models trained on WSJ and REVERB data sets together with the result of the model combination. The ST-JEM combination results when compared with the baseline combination performance show the advantage of the ST-JEM model combination for robust ASR.}
\label{tab:asr_performance}
\centering
\begin{tabular}{@{}ccccc@{}}
\toprule
\multirow{3}{*}{Test set} & \multirow{3}{*}{Model}        & \multicolumn{3}{c}{WER \%}        \\ \cmidrule(l){3-5} 
                           &                               & \multicolumn{3}{c}{Testing Model} \\ \cmidrule(l){3-5} 
                           &                               & WSJ    & REVERB   & Combination   \\ \midrule
\multirow{2}{*}{WSJ}       & \multicolumn{1}{c|}{Baseline} & 9.0    & 29.7     & 10.1          \\
                           & \multicolumn{1}{c|}{ST-JEM}     & 8.8    & 27.5     & 9.8           \\ \midrule
\multirow{2}{*}{REVERB}    & \multicolumn{1}{c|}{Baseline} & 32.0   & 8.1      & 8.3           \\
                           & \multicolumn{1}{c|}{ST-JEM}     & 31.1   & 7.5      & 7.4           \\ \bottomrule
\end{tabular}
\end{table}
Except for slight improvement in WER, we can observe very well calibrated system as shown in Figure \ref{fig:ece_speech_comb}.

\subsection{Speech JEM as multi-purpose system}\label{sec:speech_jem_future}
The general view of machine learning models is to train them on the same task that they are going to perform. We want to discuss the potential future use of JEM trained in a stable way. We suggest that JEM can model multiple joint distributions at the same time, e.g. joint distribution of inputs $\x$ and phonemes $ph$ and at the same time joint distribution of $\x$ and the identity of the $spk$. We noticed that, when we try to generate speech from the model, it has no notion of speaker identity as it was never exposed to a sequence longer than $21$ frames. Having the access to speaker identity, we can condition our generation on the fact that the likelihood of the speaker for each frame needs to be high and also the same. If the model is powerful enough to learn this complex distribution, it might function as text-to-speech (TTS). This would further allow us to perform voice conversion. Another interesting domain for JEM could be source separation, as we could follow some sort of updated version of SGLD to iteratively separate speech into two parts that sum into the original speech, where in each iteration, separated speech would be conditioned to have the high likelihood for that particular speaker. Next, the straightforward application is inpainting, we noticed that JEM is capable of very interesting results when exposed to an image of e.g. airplane and we perform SGLD conditioned on the cat class as the model is able to change the class of the image while visually on the pixel level, the resulting image is very close to the original one. This might open the possibility of more sophisticated changes than just speech conversion. 
%As we demonstrate, there are multiple possible ways to use ST-JEM. The last one that we would like to mention is connected to interpretability. We can condition our generation (SGLD procedure) on some particular settings by introducing auxiliary loss such as regularizing network activations towards desired activations. With that settings, we can try to generate input with a high likelihood. These samples can be then examined to better understand the learned properties of the network, which might be otherwise treated as uninterpretable, as this procedure would be unreliable for a discriminative model because it lacks the notion of confidence provided by the likelihood.

We found an interesting parallel between the inference through SGLD in JEM and human reasoning. Using more sophisticated inference with adaptive computation time compared to just forwarding the input through a model is more aligned with how people think. Moreover, the generative part of JEM can work as a proxy for the ability of people to be self-aware of when they do not know. This motivates future work on these models even more.

\section{Derivations}
Expressing gradient of $\log p_{\boldsymbol{\theta}}(\mathbf{x} \mid y)$ by substituting term from Equation \ref{eq:px_giv_y_intract}:
\begin{equation} \label{eq:px_giv_y_grad_full}
\nabla_{\boldsymbol{\theta}} \log p_{\boldsymbol{\theta}}(\mathbf{x} \mid y) = \nabla_{\boldsymbol{\theta}} f_{\boldsymbol{\theta}}(\mathbf{x})_y - \nabla_{\boldsymbol{\theta}} \log {\int_{\mathbf{x}} e^{f_{\boldsymbol{\theta}}(\mathbf{x})_y} d\mathbf{x}} =  \nabla_{\boldsymbol{\theta}} f_{\boldsymbol{\theta}}(\mathbf{x})_y - \E_{\mathbf{x} \sim p_{\boldsymbol{\theta}}(\mathbf{x} \mid y)} \left[\nabla_{\boldsymbol{\theta}} f_{\boldsymbol{\theta}}(\mathbf{x})_y \right]
\end{equation}
\begin{equation} \label{eq:px_giv_y_intract}
\begin{split}
\nabla_{\boldsymbol{\theta}} \log {\int_{\mathbf{x}} e^{f_{\boldsymbol{\theta}}(\mathbf{x})_y d\mathbf{x}}} & = \frac{{\int_{\mathbf{x}}  e^{f_{\boldsymbol{\theta}}(\mathbf{x})_y} \nabla_{\boldsymbol{\theta}} f_{\boldsymbol{\theta}(\mathbf{x})_y} d\mathbf{x}}}{{\int_{\mathbf{x}} e^{f_{\boldsymbol{\theta}}(\mathbf{x})_y}  d\mathbf{x}}} = \int_{\mathbf{x}} \frac{  e^{f_{\boldsymbol{\theta}}(\mathbf{x})_y}}{Z_{\boldsymbol{\theta}}} \frac{\nabla_{\boldsymbol{\theta}} f_{\boldsymbol{\theta}(\mathbf{x})_y}}{ p_{\boldsymbol{\theta}}(y)} d\mathbf{x} \\ & = \int_{\mathbf{x}} \frac{p_{\boldsymbol{\theta}}(\mathbf{x},y)}{p_{\boldsymbol{\theta}}(y)} \nabla_{\boldsymbol{\theta}}f_{\boldsymbol{\theta}(\mathbf{x})_y} = \E_{\mathbf{x} \sim p_{\boldsymbol{\theta}}(\mathbf{x} \mid y)} \left[\nabla_{\boldsymbol{\theta}} f_{\boldsymbol{\theta}}(\mathbf{x})_y \right]
\end{split}
\end{equation}

\end{document}